\documentclass[sigconf]{acmart}

\usepackage{booktabs}
\usepackage{xcolor}
\definecolor{green}{rgb}{0.0, 0.50, 0.13}
\definecolor{purple}{rgb}{0.41, 0.21, 0.61}

\AtBeginDocument{%
  \providecommand\BibTeX{{%
    \normalfont B\kern-0.5em{\scshape i\kern-0.25em b}\kern-0.8em\TeX}}}

\usepackage{tabularx}
\acmConference[ICONS '20]{ICONS 20: International Conference on Neuromorphic Systems}{June 28--30, 2018}{Chicago, IL}

\begin{document}

\title{Recurrent and Spiking Modeling of Sparse Surgical Kinematics}


\sloppy
\author{Neil Getty}
\email{ngetty@anl.gov}
\affiliation{%
  \institution{Data Science and Learning Division, \\ Argonne National Laboratory}
}

\author{Zixuan Zhao}
\email{zzhao@anl.gov}
\affiliation{%
  \institution{Data Science and Learning Division, \\ Argonne National Laboratory}
}

\author{Stephan Gruessner}
\email{sgrues2@uic.edu}
\affiliation{%
  \institution{Department of Surgery, \\ University of Illinois at Chicago}
}

\author{Liaohai Chen}
\email{lhchen@uic.edu}
\affiliation{%
  \institution{Department of Surgery, \\ University of Illinois at Chicago}
}

\author{Fangfang Xia}
\email{fangfang@anl.gov}
\affiliation{%
  \institution{Data Science and Learning Division, \\ Argonne National Laboratory}
}


\begin{abstract}
Robot-assisted minimally invasive surgery is improving surgeon performance and patient outcomes.
This innovation is also turning what has been a subjective practice into motion sequences that can be precisely measured.
A growing number of studies have used machine learning to analyze video and kinematic data captured from surgical robots.
In these studies, models are typically trained on benchmark datasets for representative surgical tasks to assess surgeon skill levels.
While they have shown that novices and experts can be accurately classified, it is not clear whether machine learning can separate highly proficient surgeons from one another, especially without video data.
In this study, we explore the possibility of using only kinematic data to predict surgeons of similar skill levels.
We focus on a new dataset created from surgical exercises on a simulation device for skill training.
A simple, efficient encoding scheme was devised to encode kinematic sequences so that they were amenable to edge learning.
We report that it is possible to identify surgical fellows receiving near perfect scores in the simulation exercises based on their motion characteristics alone.
Further, our model could be converted to a spiking neural network to train and infer on the Nengo simulation framework with no loss in accuracy.
Overall, this study suggests that building neuromorphic models from sparse motion features may be a potentially useful strategy for identifying surgeons and gestures with chips deployed on robotic systems to offer adaptive assistance during surgery and training with additional latency and privacy benefits.

\end{abstract}


\ccsdesc[500]{Computing methodologies~Machine learning}
\ccsdesc[500]{Computer systems organization~Robotics}
\ccsdesc[500]{Computer systems organization~Embedded systems}

\keywords{spiking neural network, neuromorphic, robotics, surgery, dataset}

\begin{teaserfigure}
  \includegraphics[width=\textwidth]{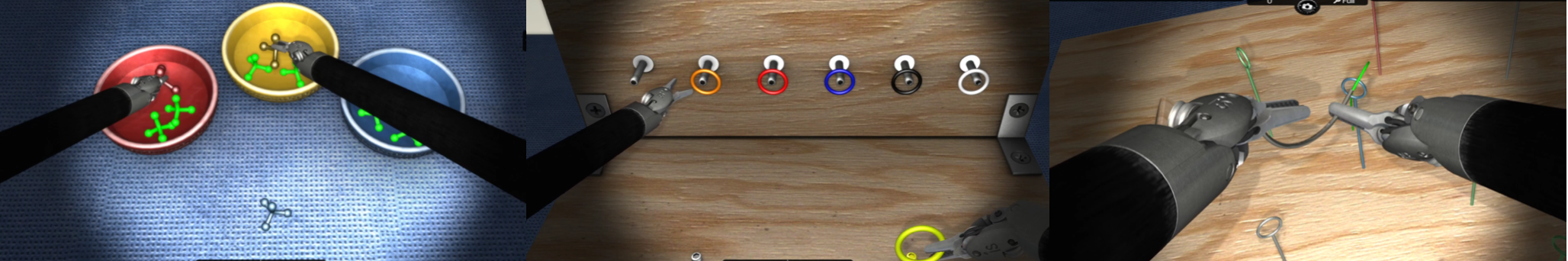}
  \caption{Robotic Surgery Training Tasks. \textmd{This article explores the possibility of predicting surgeons and surgical tasks with deep learning and neuromorphic learning. Depicted in the figure are three exercises, Pick and Place, Peg Board, and Thread the Rings, on a simulation device where surgical fellows control physical consoles that mimic surgery robots.}}
  \label{fig:teaser}
\end{teaserfigure}

\maketitle

\section{Introduction}
Robotic surgery has made tremendous progress in recent years within the medical field.
Compared with traditional practice, surgical robots nowadays enable surgeons to remotely control thin rods with intricate instruments and video cameras.
This has been applied to many types of surgery including tumor resection, microsurgical blood vessel reconstruction, and organ transplantation.
Because of the smaller incisions and decreased blood loss during the operation, patients generally experience less trauma and recover faster.
The precision, stability, and flexibility of the robotic devices also allow surgeons to perform complex procedures that would otherwise have been difficult or impossible.

The digitization of surgical vision and control has made this field ripe for the next AI revolution.
While there are many algorithmic and data challenges, the hardware is ready.
In fact, there is a push for rapid technological advancements: new robotic systems are competing to make surgeons see the targeted anatomy better and perform procedures that are ever more delicate.
At least nine surgical devices and platforms have been approved by the FDA and more are under review \cite{peters2018review}.
Accompanying this trend is the development of simulation platforms for training surgeons with similar console input, mimic exercises in virtual reality, and scoring systems for grading skills.

A clear aspiration in the field is to build autonomous surgical robots.
This however will remain out of reach for some time due to the complexity of instrument-environment interaction and soft tissue dynamics.
Most AI efforts in surgery have so far been incremental, aimed at building up our understanding of the data recorded from the robotic surgery platforms \cite{zhou2019artificial}.
A growing number of studies have shown that machine learning can help evaluate surgeon skills or delineate task structures with video and motion data.
These studies typically focus on a benchmark dataset, JIGSAWS \cite{gao2014jhu}, with three representative surgical tasks performed by eight surgeons binned into three skill levels: novice, intermediate, and expert.
The fact that these surgeons had wide practice gaps (from less than 10 hours of robotic operation time for novices to over 100 hours for experts) made the classification task well defined.

In this study, we focus on a new dataset created by surgical fellows at UIC performing training exercises with similar levels of high proficiency.
We explore the possibility of surgeon and task prediction in this challenging setting without video data.
We demonstrate that, with a simple encoding of the motion sequences, we could predict surgeon with $83\%$ accuracy and exercise with $79\%$ accuracy using kinematic data alone.
Our efficient, binary encoding of motion characteristics outperforms models on raw features and enables us to convert the models to spiking neural networks with sparse input and no performance loss.

\section{Background and Related Work}

While AI has revolutionized many fields, its impact in surgery has so far been multifaceted but incremental \cite{article}.
Medical images are processed with deep learning to help pre-operative planning and intra-operative guidance \cite{zhou2019artificial}.
Surgical videos are analyzed to assess human proficiency \cite{Funke_2019}, recognize gestures \cite{gao2020automatic, huang2018neural, sarikaya2019surgical} as well as segment surgical instruments \cite{ni2019rasnet, qin2020better} and human anatomy \cite{allan20202018, article}. 
Likewise, kinematic logs recorded from clinical surgical devices have been used in skill assessment \cite{IsmailFawaz2018evaluating, Nguyn2019SurgicalSL, zhou2015learning} and gesture recognition \cite{Krishnan2018, Sefati2015LearningS, sarikaya2020generalizable} via convolutional, recurrent, graph and neural networks. Many of these studies focused on the aforementioned JIGSAWS dataset and achieved high classification accuracy. 

There has been much effort to convert commonly used classical networks into spiking neural networks (SNN) \cite{1603.08270, 1601.04187}, in favor of its energy efficiency and biological plausibility. Recent successes in application of SNN to event-driven data \cite{blouw2018benchmarking, Stromatias2017} reveal potential in handling difference-encoded time-series data in surgery. We hypothesize that it will take advances in novel neuromorhpic hardware architectures and SNN design to enable breakthroughs in AI advanced robotic surgery.




\section{Methods}

\subsection{Data}
\label{mimic-data}
Surgical fellows at UIC have endeavoured to create a dataset of paired video and kinematic sequences recorded from surgical simulation exercises. 
These procedures were performed on MIMIC's FlexVR portable 3D standalone surgical system. 
In this study, we focus on a subset 120 exercises performed by 4 surgical fellows on 4 representative tasks training hand-eye coordination, ambidexterity, and fine motor control.
For each exercise, there are on average 4000 timesteps of kinematic data describing instrument positions, orientations, and gripper angles that correspond to 0.5-6 minute long videos. Data was stratified by both surgeon and exercise, maintain a fair class-balance and avoid any data-leak.

\subsection{Sparse encoding of kinematic sequence}

Using the MIMIC simulator dataset described in Section \ref{mimic-data}, deep learning models were trained to predict two ground truth labels  extracted from the kinematic logs. The logs begin in the format of 30 positional recordings per second describing the location and configuration of the camera and apertures. This is converted from positional data at each timestep to the difference or movement between timesteps. While the accuracy of the models trained on these movements was good, we ultimately convert these movements to binary events. All movements above a small threshold are coded as an event for that feature. While this results in an overall loss of information, the motivation is twofold; (1) the capture rate of events is sufficient that the micro movements between subsequent timesteps may be primarily operator or sensor noise and (2) encoding the data stream as binary \emph{spikes} dramatically increases the sparsity of the data, potentially greatly reducing any memory or computation footprints, and directly enabling an efficient spike-time encoding for use with neuromorphic hardware.

\subsection{LSTM and CNN models}
We trained a long short-term memory (LSTM) recurrent neural network (RNN) consisting of two bidirectional LSTM layers followed by two dense layers and final dense softmax output. For comparison, we developed convolutional neural network (CNN) with a single one dimensional convolutional layer followed by two dense layers and softmax output. The number of neurons used in each layer are [128, 64, 64, 16] and [128, 128, 16] respectively. Additional convolutional, dense layers, or increased number of neurons/filters did not improve accuracy. Dropout and batch normalization were employed between layers.

\subsection{Conversion to spiking neural networks}
The conversion from classical neural network to spiking neural network (SNN) is made with Nengo \cite{Bekolay2014}, which is a neural simulator based on the flexible Neural Engineering Framework capable of supporting multiple backends, including neuromorphic hardware such as the Intel Loihi chip \cite{loihi}, with minimal change in code. It also has a deep learning extension module NengoDL \cite{Rasmussen2018}, which is a bridge module between Keras and Nengo. Similar to other classical to spiking model conversion software like SNN Toolbox \cite{Rueckauer2017}, Nengo-DL has a converter for a deep neural network model in Keras to a native Nengo neuron model by replacing Keras layers with equivalent implementations wired with Nengo neurons. 

We used the builtin converter in NengoDL to convert the aforementioned deep neural networks to spiking neural network models. The converter uses a differentiable approximation of the non-differentiable spiking functions in neurons during the training phase and replaces it with the original function at inference \cite{hunsberger2016training}. The dense models were converted to native Nengo models without modification. The data of the convolution model was flattened as Nengo nodes only accept one-dimensional inputs. Due to the lack of recurrent support in NengoDL, the LSTM layer cannot be directly converted. As a workaround, a hybrid SNN model was created with the LSTM layer executed in Keras and the rest in Nengo.

\section{Results}

\subsection{Task and Surgeon Prediction}
\begin{table}
\caption{Model comparison for 4-class task prediction.}
\begin{tabular}{lcccc}
\toprule
\textit{Task   Model} & \textbf{LSTM}                 & \textbf{CNN} & \textbf{FCN} & \textbf{LGBM} \\ 
\midrule
\textbf{Base-Raw}     & 75.77                         & 62.89        & 63.73        & 71.57        \\
\textbf{Base-Event}   & 79.13 & 69.05        & 67.79        & 65.96        \\ 
\textbf{SNN-Raw}      & 76.71                         & 63.29        & 63.43        &          -    \\
\textbf{SNN-Event}    & \bf{79.14}                         & 67.57        & 67.00           &      -        \\
\bottomrule
\end{tabular}
\label{tab:task}
\end{table}

\begin{table}
\caption{Model comparison for 4-class surgeon prediction}
\begin{tabular}{llllc}
\toprule
\textit{Surgeon   Model} & \textbf{LSTM} & \textbf{CNN} & \textbf{FCN} & \textbf{LGBM} \\ 
\midrule
\textbf{Base-Raw}        & 80.11         & 63.32        & 60.22        & 70.31        \\ 
\textbf{Base-Event}      & 82.63         & 64.99        & 66.67        & 64.98        \\ 
\textbf{SNN-Raw}         & 75.57         & 58.71        & 60.67        &              \\ 
\textbf{SNN Event}       & \bf{83.43} & 64.71        & 65.43        &              \\ 
\bottomrule
\end{tabular}
\label{tab:surgeon}
\end{table}

\begin{figure}[!tbph]
  \centering
\includegraphics[scale=0.28]{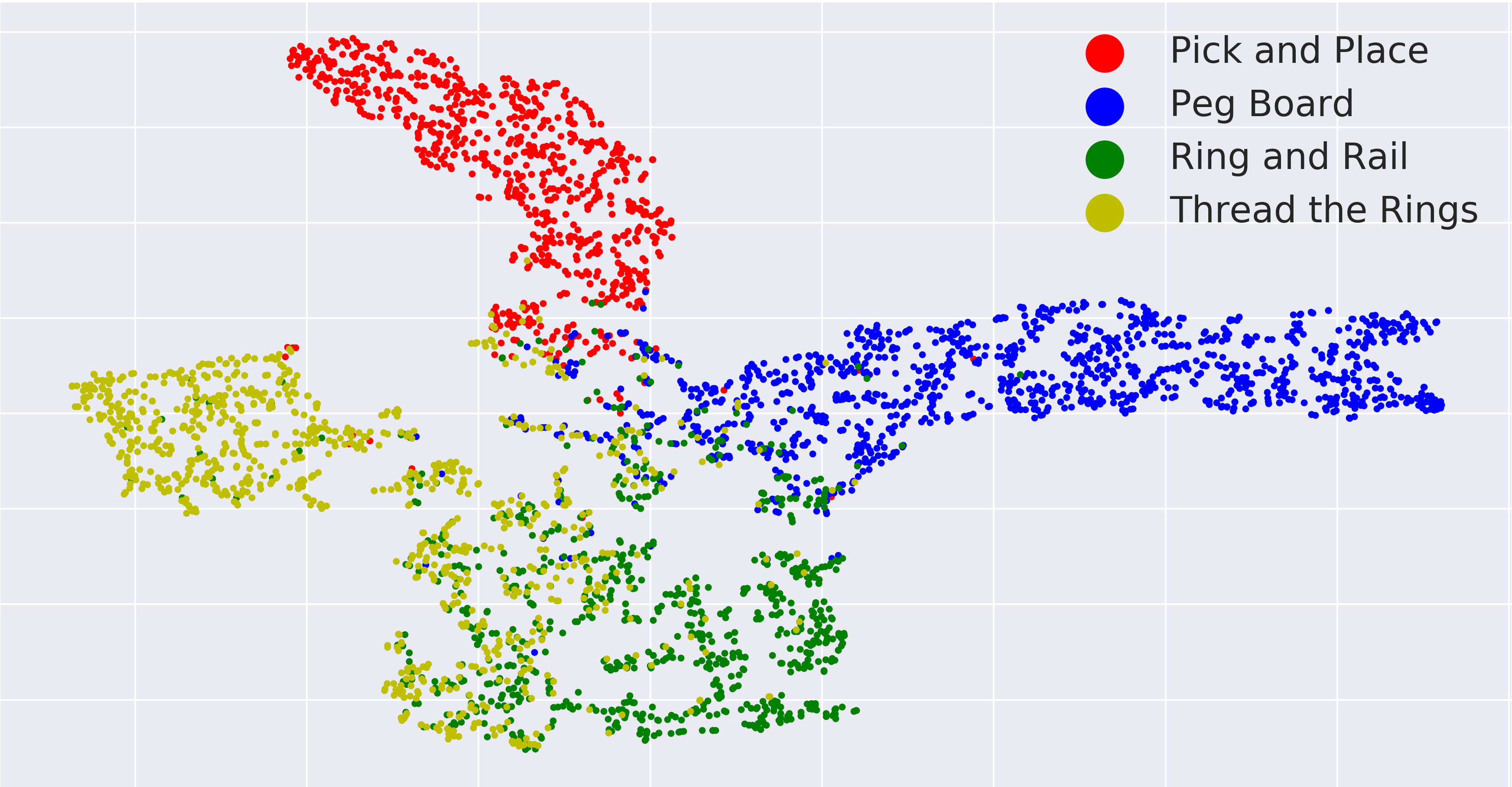}
\caption{Visualization of task based kinematic encoding}
\label{fig:task-vis}
\end{figure}

Tables \ref{tab:task} and \ref{tab:surgeon} show the comparative results of three deep learning model architectures and the decision tree method Light Gradient Boosting Machine (LGBM) \cite{lgbm}. At a high level, these results serve as a fair baseline for two easily understood tasks on our novel kinematic surgery dataset with no external data. While the sequences extracted from the kinematic data are short and may not capture  complex, long-range action associations, the recurrent neural network performs adequately at learning representations that exploit subtle differences between both surgeons and tasks. Many of these mini-action sequences should be common across many or all surgical tasks, and there are a finite many ways a surgeon can navigate and perform in a scripted environment. With more classes we should see decreased performance, but more separable and recognizable clusters of common as well as unique action sequences. Such clusters may be observed in Figures \ref{fig:task-vis} and \ref{fig:surg-vis}.

As a baseline, the best performing traditional learning algorithm (LGBM) does not outperform any deep learning methods on the event encoded data. However, the baseline model does outperform all but the LSTM on the raw kinematic motion data. For the deep learning approaches, accuracy was improved with the lossy conversion to binary event sequences. This supports our hypothesis that neuromorhpic, particularly recurrent approaches will offer an advantage for real-time event based surgical data.

\begin{figure}[!tbph]
  \centering
\includegraphics[scale=0.32]{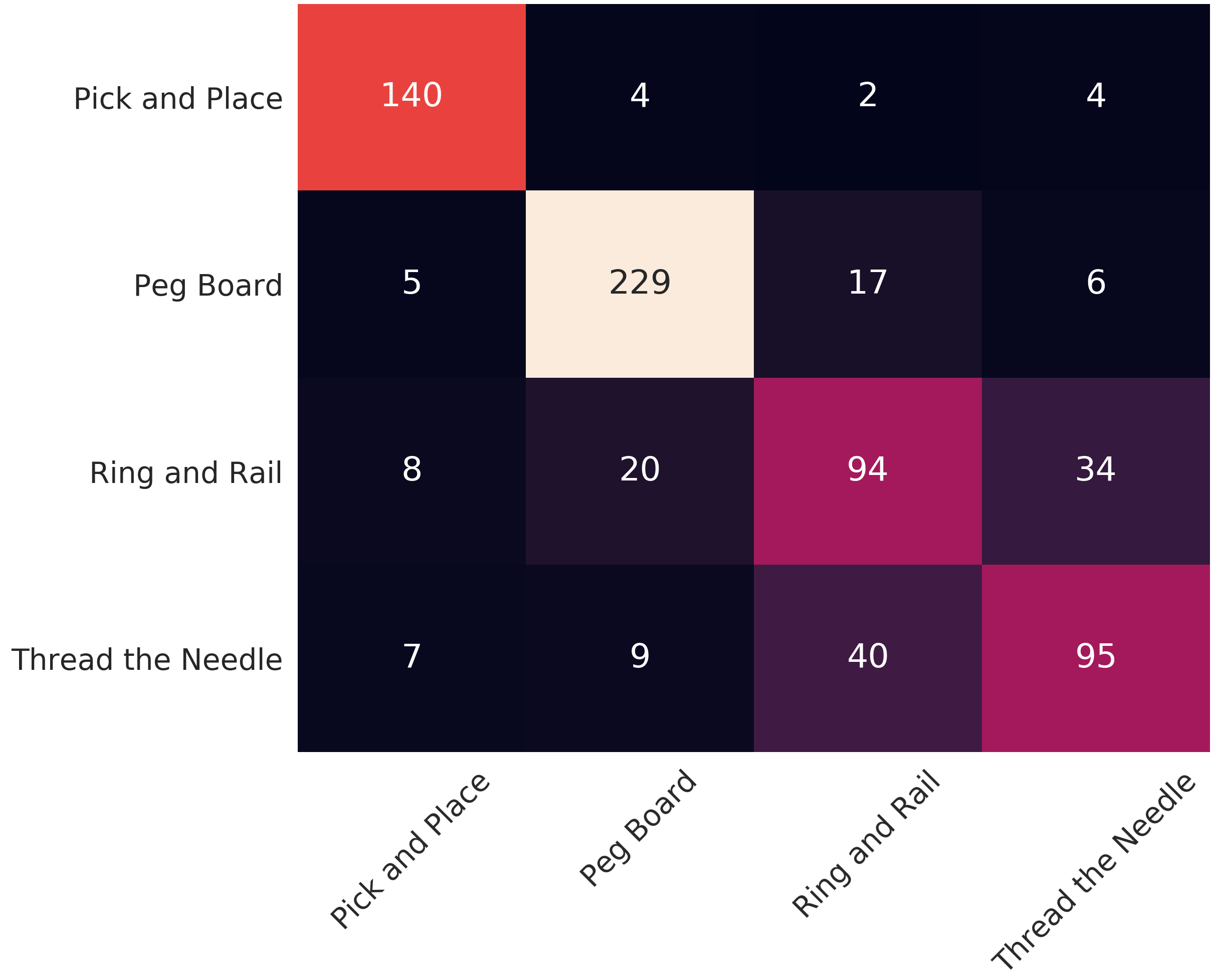}
\caption{Confusion matrix for task prediction}
\label{fig:task-conf}
\end{figure}

\begin{figure}[!tbph]
  \centering
\includegraphics[scale=0.32]{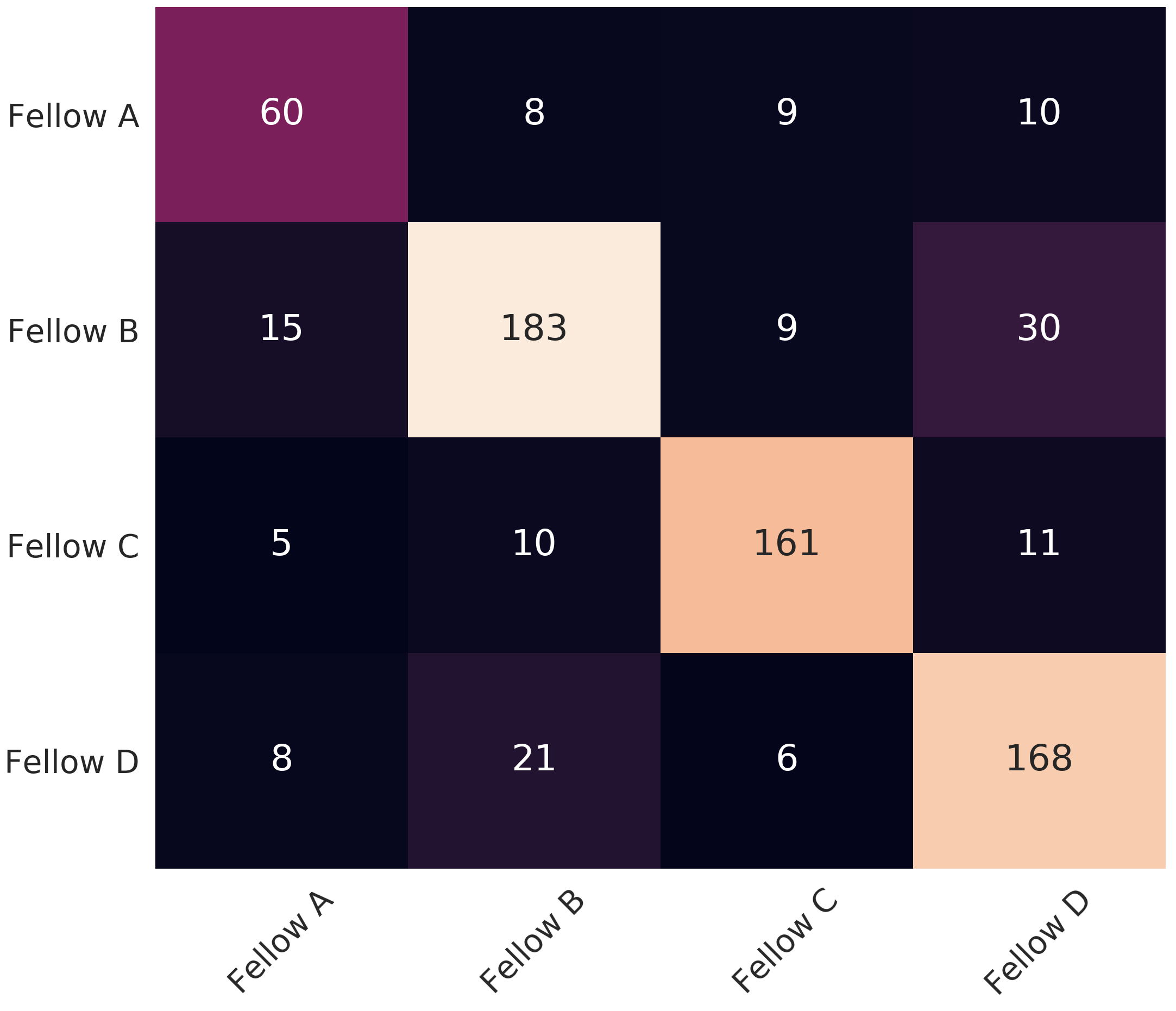}
\caption{Confusion matrix for surgeon prediction}
\label{fig:surgeon-conf}
\end{figure}

From the confusion matrix shown in Figure \ref{fig:task-conf} it is clear that most of the misclassifications are between the Ring \& Rail and Thread the Rings tasks. These tasks frequently require rotating of the wrists, direct hand offs of objects, and manually moving the camera. The Pick and Place task does not require any of these key movements. Note that while the data was stratified by surgeon and exercise the total number of sequences varies slightly due to completion time.

\subsection{Visualization of kinematic actions}

Following the input encoding and model training in the previous sections, visualizations of the latent-layer kinematic actions are generated as follows:

\begin{itemize}
\item The 20 dimensional, 40 step time-series kinematic sequences are propagated through the network until the final 16 neuron hidden Dense layer. This is the layer immediately preceding the output of the model.
\item The 16 dimensional vectors are decomposed to two dimensions with t-distributed stochastic neighbor embedding (t-SNE) \cite{vanDerMaaten2008} an approach similar to principle components analysis (PCA).
\item These 2D coordinates are plotted and colored according to the ground truth label used to train the model.
\end{itemize}

\begin{figure}[!tbph]
  \centering
\includegraphics[scale=0.28]{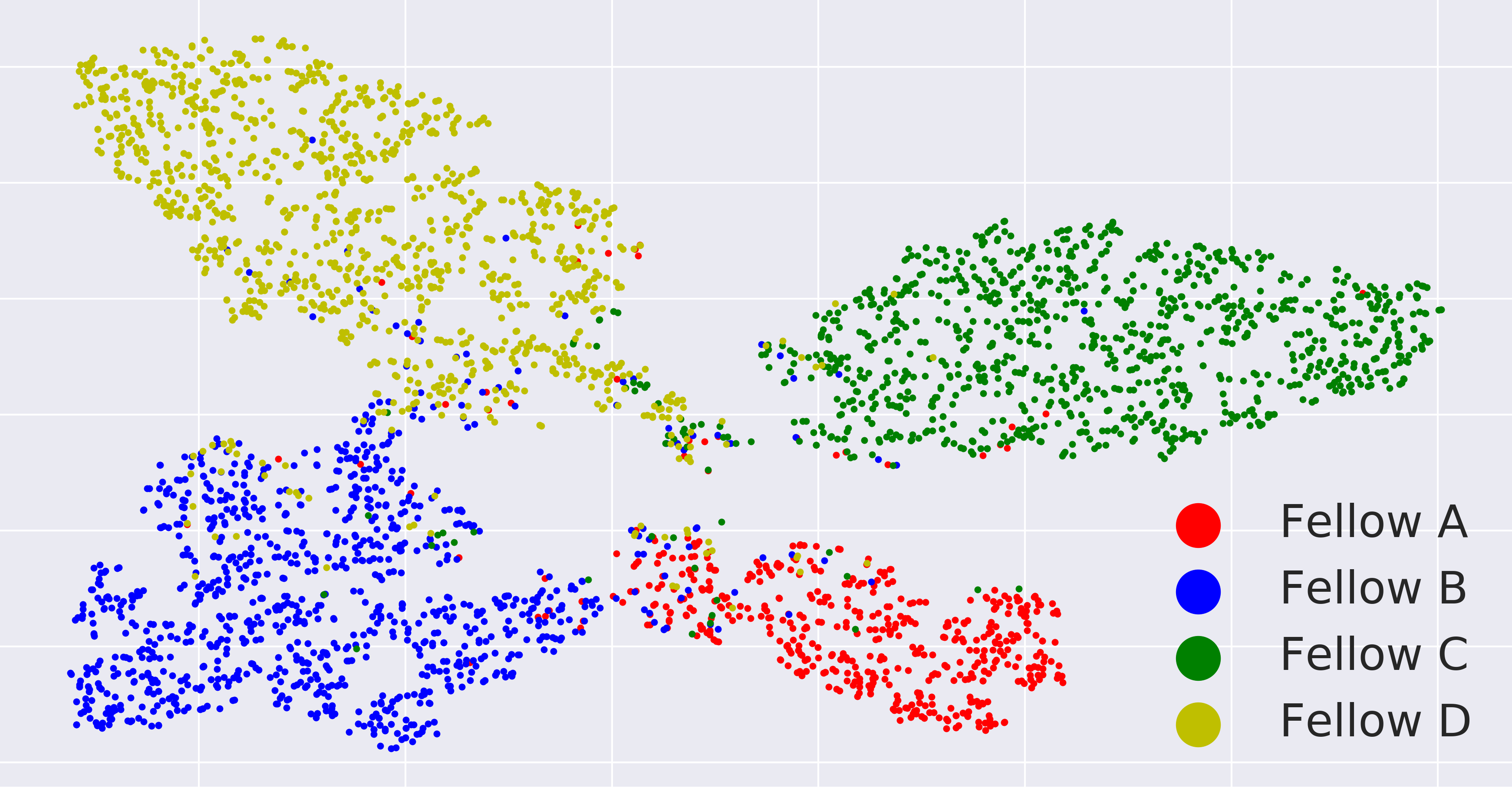}
\caption{Visualization of surgeon based kinematic encoding}
\label{fig:surg-vis}
\end{figure}

We may observe in 2D the separability of the classes, as well as clusters of similar actions across 2 or more classes. This is particularly noticeable for the surgeon prediction visualization shown in figure \ref{fig:surg-vis} which has a noticeable cluster for each permutation of the surgical fellows. Also of note is the difference in spread of the surgeon's movements. Fellow A's actions are fairly tightly distributed, while Fellow C and D's movements are spread farther apart despite having completed the same surgical tasks. 
The fact that our model appears to capture motion signatures of different surgical fellows of similar proficiency is promising. This indicates neuromorphic chips deployed on edge could be potentially useful in providing personalized assistance in surgery and training.

\subsection{Neuromorphic approach}

We compared the performance of the base and converted SNN models with the binary-encoded dataset. As shown in Table \ref{tab:task} and \ref{tab:surgeon}, the best-performing model is the LSTM, which achieves 79\% test accuracy for surgery task prediction and 83\% for surgeon prediction. Meanwhile, the convolution neural network model and fully connected model have roughly 10\% and 7\% accuracy drain respectively. Among the converted models, both the hybrid LSTM model and the other native Nengo models have roughly achieved the same accuracy as their DNN counterparts, suggesting no noticeable loss of performance in the spiking neural network conversion.

\begin{figure}
  \includegraphics[width=0.95\linewidth]{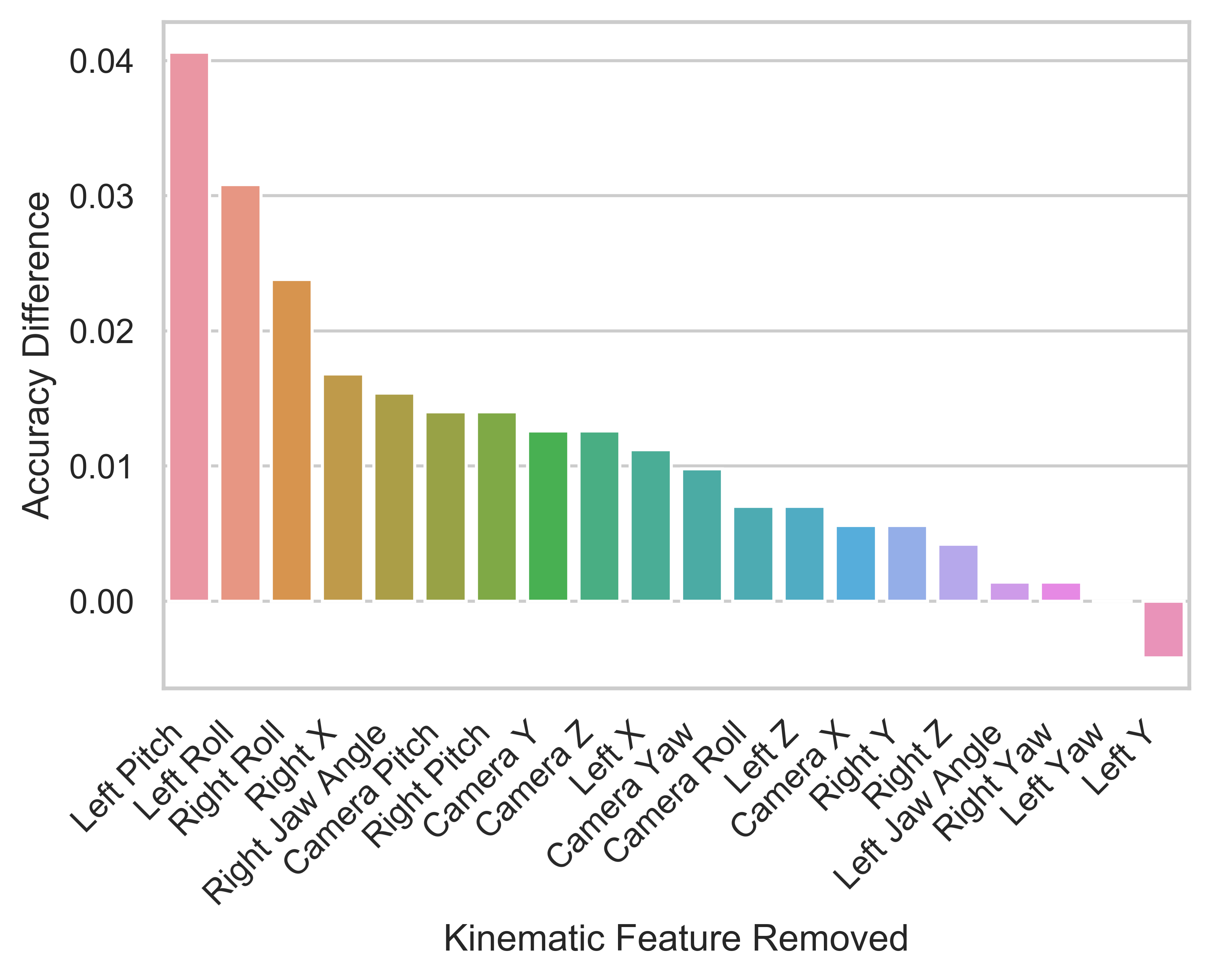}
  \caption{Feature Importance for Surgeon Prediction. \textmd{Bars represent accuracy differences between an LSTM model trained on all features and the same model with one of the features removed. }}
  \label{fig:feature_importance}
\end{figure}
\subsection{Feature Importance}
To identify the kinematic features that distinguish different surgeons, we examined the contribution of each feature to prediction accuracy. A traditional method is to query machine learning models directly for feature importance. However, this method would give the importance of a feature instance at some specific time stamp instead of its influence over the whole time series. To address this issue, we repeatedly removed one feature from the input data and ran the top performing LSTM models with identical configurations. The resulting accuracy changes are shown in Fig. \ref{fig:feature_importance} for all single features. 
The results show that certain features play an essential role in surgeon and task prediction. For example, \textit{Pitch} and \textit{Roll} actions on the left arm are top features for both surgeon and task classification. This could be explained by the nature of the exercises: \textit{Pick and Place} requires almost no rotation, while in \textit{Thread the Rings}, the surgeons tend to pick, rotate and pierce the needle through the ring with the left arm. 
There appears to be no standout feature that have an oversized impact, which again suggests that the model has learned representations of motion characteristics from the sequence of actions.

\section{Future work}
We are excited by recent advancements in neuromorphic hardware and algorithm design and applicability to open challenges in robotic surgery. Given our success with RNNs, on short sequences, we are keen to explore Applied Brain Research's LMU \cite{voelker2019lmu} networks for long-term memory associations in very long sequences. Additionally, we must explore the potential of models and techniques for domain adaptation, leveraging data from programmable simulators for clinical application. Following the success of our event encoding scheme for this novel time-series kinematic data set and conversion to SNN with no accuracy loss, we must investigate the computation speed and other advantages of real-time on-chip processing for surgical tasks using neuromorphic hardware.

\begin{acks}
Part of this work is supported by the Laboratory Directed Research \& Development Program at Argonne National Laboratory. 
We thank the surgical fellows, Dr. Alberto Mangano, Dr. Gabriela Aguiluz, Dr. Roberto Bustos, Dr. Valentina Valle and Dr. Yevhen Pavelko from Dr. Pier Cristoforo Giulianotti's robotic surgery team, for helping create and continuing to expand this dataset.
We would also like to show gratitude to Mimic Technologies Inc. for loaning the FlexVR simulation device.
We are also grateful to Nathan Wycoff for discussion and valuable comments on the manuscript. This material is based upon work supported by Laboratory Directed Research and Development (LDRD) funding from Argonne National Laboratory, provided by the Director, Office of Science, of the U.S. Department of Energy under Contract No. DE-AC02-06CH11357".
\end{acks}

\bibliographystyle{ACM-Reference-Format}
\bibliography{biblio}

\appendix

\end{document}